\documentclass[10pt,twocolumn,letterpaper]{article}

\usepackage[final]{cvpr}      

\usepackage{graphicx}
\usepackage{amsmath}
\usepackage{amssymb}
\usepackage{pifont}
\usepackage{booktabs}
\usepackage{multicol}
\usepackage{multirow}
\usepackage{tabularx}

\usepackage[pagebackref,breaklinks,colorlinks]{hyperref}

\usepackage[capitalize]{cleveref}
\crefname{section}{Sec.}{Secs.}
\Crefname{section}{Section}{Sections}
\Crefname{table}{Table}{Tables}
\crefname{table}{Tab.}{Tabs.}

\usepackage{xcolor}
\usepackage{float}
\usepackage{lipsum} 
\usepackage{boldline} 
\usepackage{colortbl}
\usepackage{wrapfig}
\usepackage{soul}
\usepackage{cleveref}

\definecolor{lightgray}{gray}{0.9}

\newcommand{\OURS}{MM-Narrator}
\newcommand{\OURSZS}{MM-Narrator$_{\mathcal{ZS}}$}
\newcommand{\MMICL}{MM-ICL}
\newcommand{\MADEVAL}{MAD-eval-\texttt{Named}}
\newcommand{\MADTRAIN}{MAD-v2-\texttt{Named}}
\definecolor{teaser_blue}{RGB}{162,193,207}
\definecolor{teaser_green}{RGB}{71,211,90}
\definecolor{teaser_pink}{RGB}{216,110,204}
\definecolor{teaser_yellow}{RGB}{205,200,0}
\definecolor{teaser_orange}{RGB}{233,113,50}

\newcommand{\okmark}{{\textbf{\textcolor[rgb]{0.1, 0.5, 0.1}{$\checkmark$}}}}
\newcommand{\ngmark}{{\textbf{\color{red}{\ding{55}}}}}

\usepackage[font=small,labelfont=bf,skip=2pt]{caption}
\newcommand{\tablestyle}[2]{\setlength{\tabcolsep}{#1}\renewcommand{\arraystretch}{#2}\centering\footnotesize}

\usepackage{ifthen}
\usepackage{textcomp}
\usepackage{etoolbox}
\usepackage{comment}
\newcommand{\Paragraph}[1]{\vspace{0mm} \noindent \textbf{#1} \hspace{0mm}}

\newcommand{\specialfootnote}[1]{%
    \begingroup%
    \renewcommand{\thefootnote}{*}%
    \footnotetext{#1}%
    \endgroup%
}

\def\cvprPaperID{6101} 
\def\confName{CVPR}
\def\confYear{2024}

\begin{document}

\title{
\begin{minipage}{\linewidth}
    \begin{minipage}[c]{0.01\linewidth}
        \centering
        \includegraphics[scale=0.08]{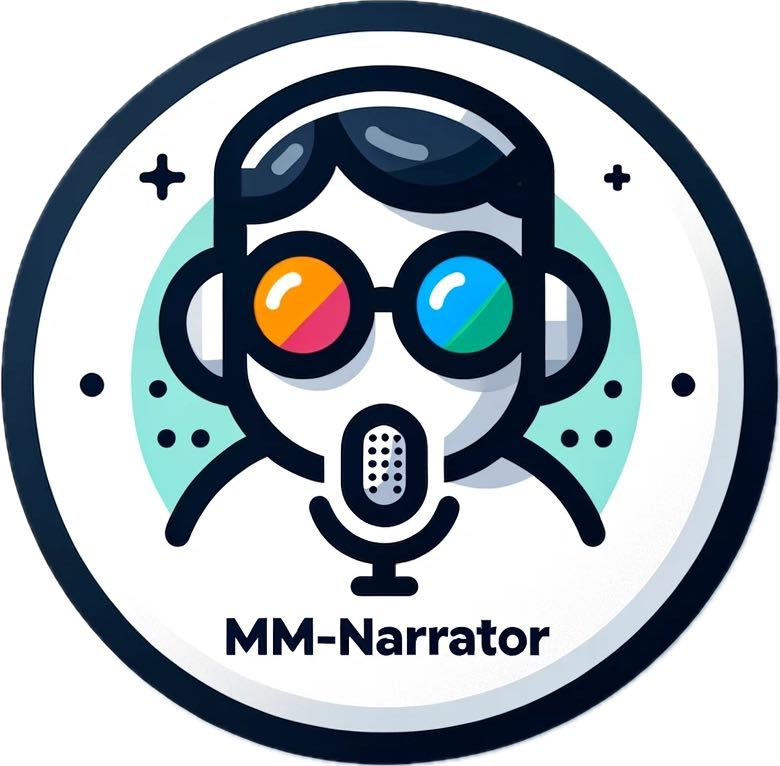} 
    \end{minipage}
    \begin{minipage}[c]{0.99\linewidth} 
      \centering
      \OURS: Narrating Long-form Videos with\\ Multimodal In-Context Learning\\

    \end{minipage}
\end{minipage}
}
\author{%
    Chaoyi Zhang$^{\spadesuit}$\footnotemark \and
    Kevin Lin$^{\clubsuit}$\and
    Zhengyuan Yang$^{\clubsuit}$  \and
    Jianfeng Wang$^{\clubsuit}$ \and
    Linjie Li$^{\clubsuit}$ \and
    Chung-Ching Lin$^{\clubsuit}$ \and
    Zicheng Liu$^{\clubsuit}$ \and
    Lijuan Wang$^{\clubsuit}$
}

\newcommand{\affiliations}{
    \begin{center}
        \vspace{-0.85cm}
         $^{\spadesuit} $University of Sydney $^{\clubsuit }$Microsoft Azure AI\\
        \href{https://MM-Narrator.github.io}{https://MM-Narrator.github.io}
    \end{center}
}

\twocolumn[{%
	\renewcommand\twocolumn[1][]{#1}%
	\maketitle
        \affiliations
	\begin{center}
	\includegraphics[width=1.0\linewidth]{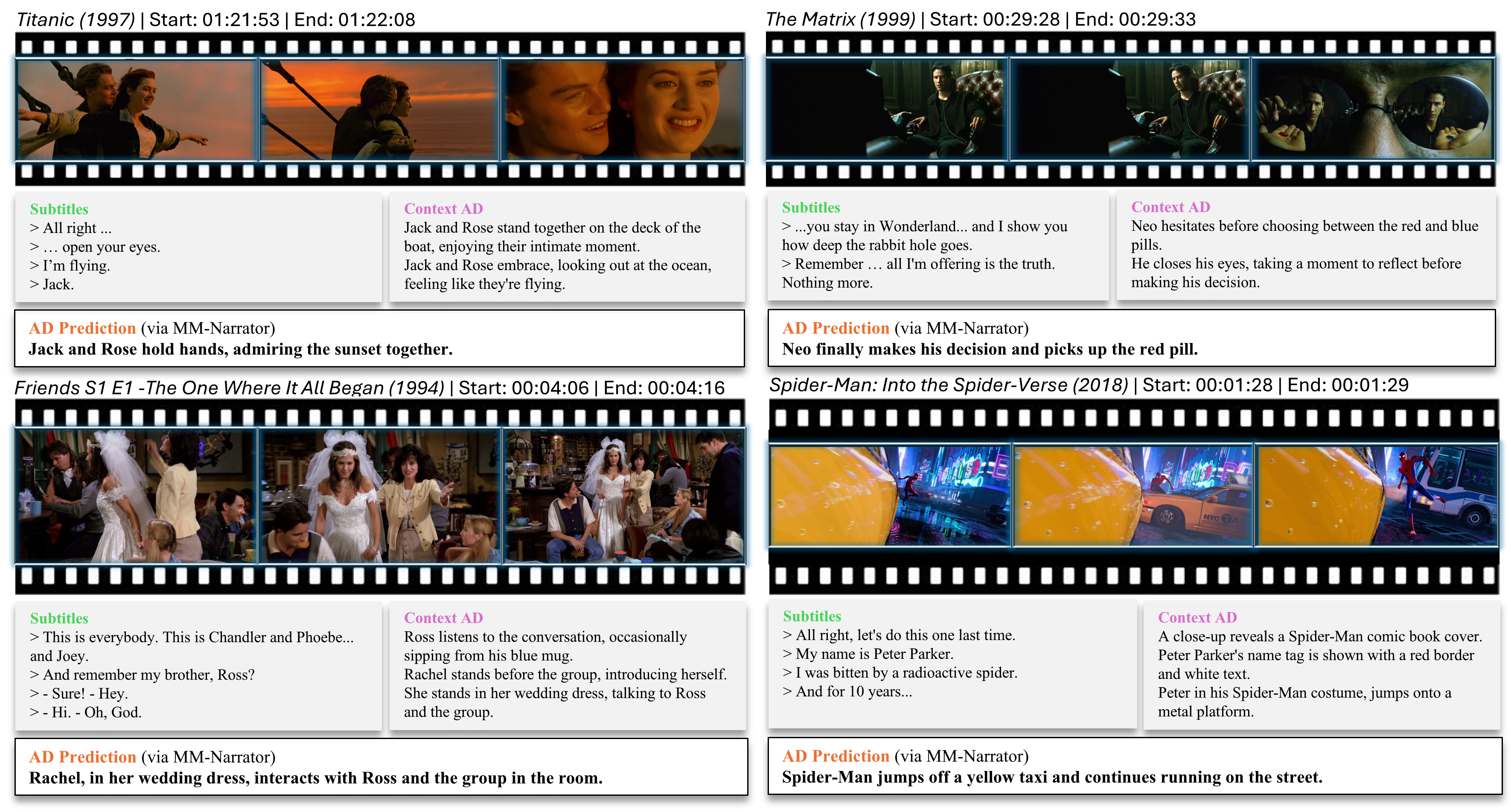} 
\captionof{figure}{We present \OURS{}, a training-free framework towards automatic audio description (AD) generation for long-form videos via iterations: for each scene, it perceives multimodal inputs (i.e., seeing \textcolor{teaser_blue}{\textbf{visual frames}} and hearing \textcolor{teaser_green}{\textbf{character dialogues}}), recalls the \textcolor{teaser_pink}{\textbf{context AD}} depicting past scenes, and infers \textcolor{teaser_orange}{\textbf{AD prediction}} for the current scene. Zoom in for details.
}
		\label{fig:teaser}
	\end{center}
}
]

\maketitle
\specialfootnote{work done during internship at Microsoft}

\begin{abstract}

We present \OURS, a novel system leveraging GPT-4 with multimodal in-context learning for the generation of audio descriptions (AD). Unlike previous methods that primarily focused on downstream fine-tuning with short video clips, \OURS~excels in generating precise audio descriptions for videos of extensive lengths, even beyond hours, in an autoregressive manner. 
This capability is made possible by the proposed memory-augmented generation process, 
which effectively utilizes both the short-term textual context and long-term visual memory through an efficient register-and-recall mechanism. 
These contextual memories compile pertinent past information, including storylines and character identities, ensuring an accurate tracking and depicting of story-coherent and character-centric audio descriptions. 
Maintaining the training-free design of \OURS{}, we further propose a complexity-based demonstration selection strategy to largely enhance its multi-step reasoning capability via few-shot multimodal in-context learning (MM-ICL).
Experimental results on MAD-eval dataset demonstrate that \OURS~consistently outperforms both the existing fine-tuning-based approaches and LLM-based approaches in most scenarios, as measured by standard evaluation metrics. 
Additionally, we introduce the first segment-based evaluator for recurrent text generation. Empowered by GPT-4, this evaluator comprehensively reasons and marks AD generation performance in various extendable dimensions.


\end{abstract}
\section{Introduction}
\label{sec:intro}

Audio Description (AD) is an essential task that transforms visual content into spoken narratives~\cite{AD}, primarily assisting visual impairments in accessing video content. 
Given its evident importance, the notable expectations for AD to fulfill include complementing the existing audio dialogue, enhancing viewer understanding, and avoiding overlap with the original audio.
This process involves identifying not just who is present in the scene and what actions are taking place, but also precisely how and when the actions occur. 
Additionally, AD should capture subtle nuances and visual cues across different scenes, adding layers of complexity to its generation.

In addition to aiding visually impaired audiences, AD also enhances media comprehension for autistic individuals, supports eyes-free activities, facilitates child language development, and mitigates inattentional blindness for sighted users~\cite{perego2016gains, webpage_ad4all}.
However, traditional human-annotated AD, while detailed, incurs significant costs and often suffers from inconsistencies due to low inter-annotator agreement~\cite{han2023autoad2}, highlighting the need for automatic AD generation systems.
Furthermore, AD serves as an emerging testbed for benchmarking the capabilities of LLM/LMM systems in long-form multimodal reasoning~\cite{han2023autoad1, han2023autoad2, 2023mmvid}, towards next-level advanced video understanding.


In this paper, we present \OURS{}, a multimodal AD narrator, to effectively leverage multimodal clues, including visual, textual, and auditory elements, to enable comprehensive perception and reasoning. 
In particular, \OURS{} distinguishes itself by naturally identifying characters through their dialogues, in contrast to existing methods that may underutilize subtitles~\cite{han2023autoad1,han2023autoad2}. 

Apart from an intricate multimodal understanding of the video content, generating story-coherent AD for long-form videos also relies on an accurate tracking and depicting of character-centric evolving storylines over extended durations, even spanning hours.
This differs AD generation from conventional dense video captioning~\cite{Wang_2019_ICCV,xu2016msr,krishna2017dense,li2021value}: 
Unlike mere frame-by-frame scene description, AD should weave a coherent narrative, utilizing characters as pivotal elements to maintain an uninterrupted storytelling flow~\cite{AD}. 
To achieve contextual understanding, we propose to leverage both short-term and long-term memories to assist \OURS{} in its recurrent AD generation process.
Specifically, short-term textual memory sets the stage for generating coherent narrations, whereas long-term visual memory aids in character re-identification during long-form videos, especially for scenes lacking dialogue.

As a GPT-4 empowered multimodal agent, \OURS{} could further benefit from multimodal in-context learning (MM-ICL) via our proposed complexity-based multimodal demonstration selection. 
With complexity defined with the chain-of-thought (CoT) technique~\cite{wei2022chain}, \OURS{} could efficiently form and learn from a smaller candidate pool of multimodal demonstrations, effectively improving its multimodal reasoning capability in a few-shot approach.
This proposed complexity-based selection surpasses both random sampling and similarity-based retrieval, which are classic ICL solutions in choosing few-shot examples.

In summary, our contributions are four-folds:
\textit{(1)}~
We present \OURS{}, an automatic AD narrator for long-form videos that can perceive multimodal inputs, recall past memories, and prompt GPT-4 to produce story-coherent and character-centric AD.
\textit{(2)} We propose a complexity-based multimodal in-context learning (MM-ICL) to further boost its AD generation performance with few-shot examples, offering new insights into the question \textit{``what makes good ICL examples?"} under complex text generation scenarios with multimodal reasoning needed. 
\textit{(3)} Our training-free \OURS{} outperforms both fine-tuning-based SOTAs and LLM/LMM baselines, including GPT-4V, in most classic captioning metrics. 
\textit{(4)} Furthermore, we introduce the first GPT-4 based evaluator for recurrent text generation, measuring more comprehensive AD generation qualities at both text-level and sequence-level. 
Results suggest that \OURS{} generates AD comparable to human annotations across several considered aspects.


\section{Related Work} 



\begin{figure*}[th] 
    \centering
    \includegraphics[width=\linewidth]{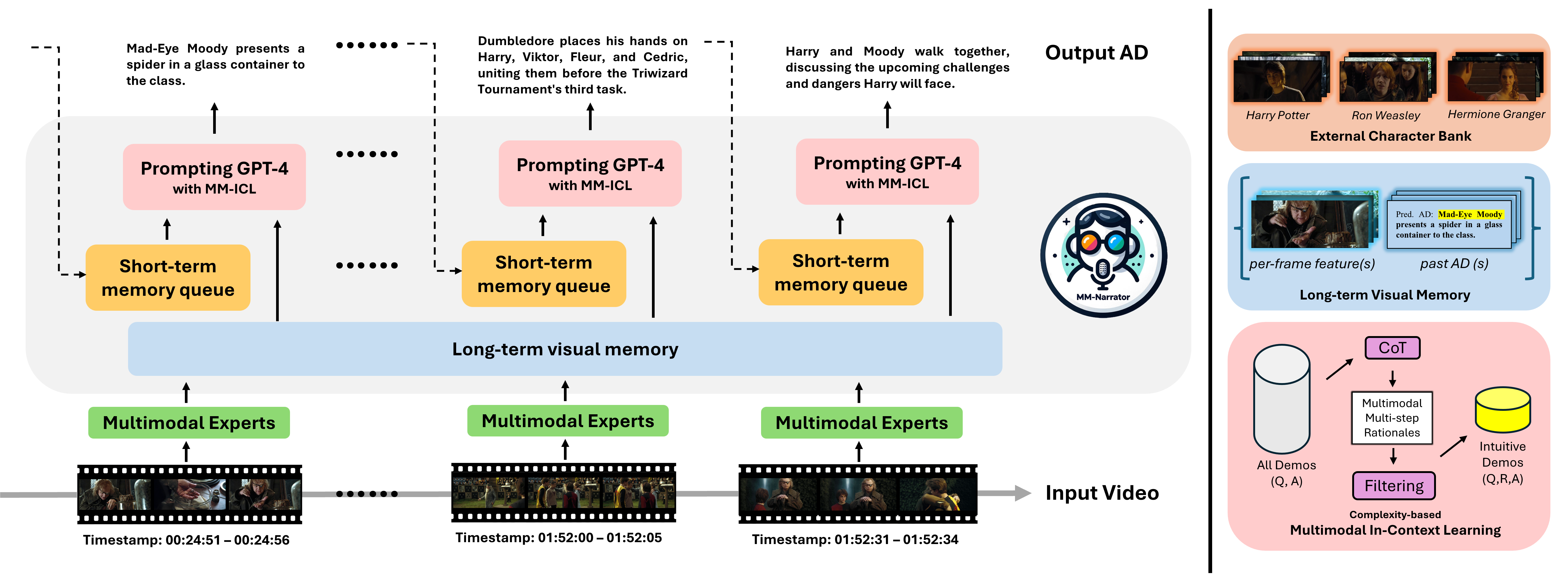}
    \caption{\OURS{} generates AD sequence via iterations. }
    \vspace{-0.35cm}
    \label{fig:overall}
\end{figure*}

\Paragraph{Audio Description} (AD)
offers verbal narration of key visual elements in videos~\cite{AD}, enriching the viewing experience for individuals who are blind or have low vision. AD differs from video captioning~\cite{aafaq2019video,Wang_2019_ICCV,xu2016msr,krishna2017dense,li2021value,lin2022swinbert}, which solely describes the visual content of a given video clip. Instead, AD generation considers multiple modalities, aiming to generate coherent narratives of storylines, characters, and actions in a way that complements the regular audio track. Initial studies~\cite{rohrbach2015dataset,rohrbach2017movie,torabi2015using,soldan2022mad} concentrated on developing audio segmentation and transcription system to collect high-quality video datasets with temporally aligned ADs. These foundational efforts pave the way for more advanced explorations in LSMDC~\cite{rohrbach2017movie}. Recent research~\cite{han2023autoad1} has ventured into training transformer models equipped with a frozen LLM. Researchers also incorporate an external character bank~\cite{han2023autoad2} to enhance the accuracy of AD generation. Different from prior works~\cite{han2023autoad1,han2023autoad2} that rely on downstream fine-tuning, our proposed \OURS~generates accurate ADs in a training-free manner.
\\

\Paragraph{LLM for Video Understanding.} The remarkable success of Large Language Models (LLMs)~\cite{gpt4,chowdhery2022palm,touvron2023llama,vicuna2023,alayrac2022flamingo,driess2023palm} has sparked increasing interest in their application to video understanding. Recent works~\cite{VLog,bhattacharya2023video,2023mmvid,luo2023valley,song2023moviechat,2023videochat} generally fall into two main categories: (\textit{i}) visual instruction tuning, and (\textit{ii}) prompting LLMs. The first approach~\cite{liu2023llava,maaz2023video,2023videochat,luo2023valley} typically fine-tunes an LLM-based model. This involves integrating the pre-trained LLMs and additional trainable networks. 
The second category~\cite{langchain} involves prompting LLMs to invoke specialized expert tools, transforming the input video into a textual document, which then serves as input to the LLMs for reasoning~\cite{VLog,bhattacharya2023video,2023mmvid}. However, this strategy may not be effective for processing lengthy or speech-dense videos, as the LLMs often face challenges with excessive token lengths. Different from prior work, we propose to leverage short-term textual memory and long-term visual memory with a register-and-recall mechanism, to effectively generate ADs for long-form videos.

\Paragraph{In-Context Learning}(ICL)~\cite{brown2020language, min2021metaicl,min2022rethinking, liu2023pre, lu2022learn}, as a new paradigm, allows LLMs to learn from a few examples without needing parameter updates via downstream fine-tuning.
This learning-from-analogy strategy~\cite{dong2022survey} augments original query question with a context formed by natural language demonstrations.
Existing studies highlight that the success of ICL largely depends on the selection of effective demonstrations.
One common solution ~\cite{liu2021makes,rubin2021learning} is to form the ICL prompt with closest neighbors, which are retrieved with highest similarity to the query embedding.
Other query-based metrics are also explored in finding supportive ICL examples on the basis of query content, such as mutual information~\cite{tanwar2023multilingual,sorensen2022information} and perplexity~\cite{gonen2022demystifying}.
Although prior works have demonstrated their superiority in text classification tasks or open-domain QA~\cite{dong2022survey}, they have not explored ICL on complex text generation tasks under multimodal scenarios.
In this work, we propose to quantify the demonstration complexity as the number of reasoning steps in chain-of-thoughts (CoTs)~\cite{wei2022chain,zhang2022automatic}, and select the most intuitive examples to improve AD generation with few-shot MM-ICL.




\section{Method}



Given a long-form video $\mathcal{V}$, consisting of multiple video clips ${\{\boldsymbol{v}_t\}}$, \OURS{} generates an AD sequence $\{\mathcal{T}_t\}$ in an autoregressive manner, as shown in Figure ~\ref{fig:overall}.
We first present \OURS{}, a multimodal narrator that conducts recurrent AD generation via prompting GPT-4 (\S\ref{subsec:zeroshot}).
Building upon \OURS{}, we propose the complexity-based MM-ICL to further enhance its multimodal reasoning capabilities through intuitive few-shot demonstrations (\S\ref{subsec:mmicl}). 
Notably, the entire \OURS{} framework operates in a training-free manner.


\subsection{Recurrent AD Narrator}
\label{subsec:zeroshot}

At each iteration of scanning through a specific long-form video, \OURS{} utilizes multimodal experts for perception, recalls past memories in both short-term and long-term contexts, and prompts LLM to generate an audio description. We describe each step as below.


\Paragraph{Multimodal perceptions.} 
We employ specialized vision and audio expert models to extract multimodal information from the input video clip.
We denote a video clip consisting of $N$ frames with timestamp $t$ as $\boldsymbol{v}_t= \{ \mathcal{I}_1, \mathcal{I}_2, ...,\mathcal{I}_N\} $.
We deploy vision experts~\cite{radford2021learning,wu2022grit,wang2023yolov7} to gather visual perceptions, which involves obtaining per-frame visual features and text-formed outputs. 
Specifically, for each frame $\mathcal{I}_i$, we collect CLIP-ViT features $\boldsymbol{x}^{CLIP}_{i}$, image captions $\boldsymbol{x}^{cap}_{i}$, and people detections $\boldsymbol{x}^{det}_{i}$.
Alongside these crucial visuals, we observe the spoken dialogues play a profound role, which is underutilized in existing approaches~\cite{han2023autoad1,han2023autoad2}.
The spoken dialogues not only offer information complementary to the visuals, but also primarily serve as the only access to identify characters with their names when no external video metadata is given.
To be specific, we concatenate the subtitles within a certain time window $T_{sub}$ as $\boldsymbol{x}^{sub}_{t\in T_{sub}}$.
These subtitles can be sourced from the Internet or generated through automated speech recognition (ASR) as an audio expert~\cite{bain2022whisperx}.
To summarize, for a given video clip $\boldsymbol{v}_t$, the multimodal experts produce a comprehensive tuple of perception clues 
$\mathcal{X}_t = (
\{\boldsymbol{x}^{CLIP}_{i}\}, 
\{\boldsymbol{x}^{cap}_{i}\}, 
\{\boldsymbol{x}^{det}_{i}\}, 
\boldsymbol{x}^{sub}_{t\in T_{sub}}
)$, where $\{\boldsymbol{x^{\cdot}_i}\}$ denotes the per-frame outputs. Among these, $\{\boldsymbol{x}^{cap}_{i}\}$ and $\boldsymbol{x}^{sub}_{t\in T_{sub}}$ are directly used in constructing LLM prompts, while the others facilitate the proposed register-and-recall mechanism for long-term character re-identification.



\Paragraph{Short-term memory queue.} To equip \OURS{} with contextual understanding for coherent AD generation, we maintain a short-term memory queue $\mathcal{M}_{\boldsymbol{short}} = \{\mathcal{T}_{t-K}, ..., \mathcal{T}_{t-1}\}$ to contain the $K$ most recently predicted ADs with timestamps. 
The short-term memory queue will be updated over time during inference.
This lightweight textual queue is instrumental in creating story-coherent AD narrations, enabling visually impaired audiences to follow the storytelling more intuitively.


\Paragraph{Long-term visual memory.} To endow \OURS{} with the ability to recall characters identified in previous video clips, we construct a frame-level character re-identification visual bank.
This visual bank, designed for long-term use, is operated by a register-and-recall mechanism as follows: (1) we register $\boldsymbol{x}^{CLIP}_{j}$ as the visual signature for each globally-indexed frame $\mathcal{I}_j$ in all previous video clips $ \mathcal{I}_j \in \{ v_1, v_2, ..., v_{t-1}\}$, and (2) for each current frame $\mathcal{I}_i$, we first filter-out the invalid matches resulting in nonpositive cosine similarity $\mathtt{Sim}_{cos}(\boldsymbol{x}^{CLIP}_{i},\boldsymbol{x}^{CLIP}_{j})$, and then retrieve the past predicted AD which owns the highest similarity to the current visual signature $\boldsymbol{x}^{CLIP}_{i}$.
For simplicity, this mechanism is activated only when a single individual is detected in a frame (i.e., $|\boldsymbol{x}^{det}_{\cdot}|=1$), typically in close-up shots of the character, making frame-level CLIP-ViT features~\cite{radford2021learning} compatible for character re-identification.
Additionally, the retrieval candidate pool is refined to include only past predicted ADs where person named entities are recognized through a Named Entity Recognition (NER) tool~\cite{finkel2005incorporating}.
This strategy focuses \OURS{} on the main characters who contribute to the past storyline.



\Paragraph{Prompting LLM for AD generation.}
Gathering all aforementioned text-formed outputs, \OURS{} builds prompts to query GPT-4 for recurrent AD generation. 
Specifically, the input prompt contains the following elements: task introduction, visual captions  ($\boldsymbol{x}^{cap}_{i}$) with successfully re-identified characters, recent context ADs ($\mathcal{M}_{\boldsymbol{short}}$) and character dialogues ($\boldsymbol{x}^{sub}_{t\in T_{sub}}$).
Noticeably, we also found that adding task-specific hints into the prompt could empirically benefit overall AD generation, which we attribute as an explicit attention guidance via prompt engineering. 
A breakdown of our AD generation prompt constructed by \OURS{}, is provided in the appendix (Figure~\ref{fig:breakdown}).



\subsection{Multimodal In-Context Learning}
\label{subsec:mmicl}
In this section, we further extend \OURS{} with multimodal in-context learning (MM-ICL) on few-shot examples.
Our exploration begins by examining two primary methods of demonstration selection: random and similarity-based approaches. We then critically evaluate the question, ``\textit{What makes for effective ICL examples?}'' and propose a complexity-based MM-ICL approach to improve the multimodal reasoning capability with the most intuitive multimodal demonstrations.

\Paragraph{Random MM-ICL.} Firstly, we build an in-context learning (ICL) demonstration pool, denoted as $\mathcal{P}$, from the training dataset. Each demonstration within the pool is composed of a pair $(\mathcal{Q}, \mathcal{A})$, where $\mathcal{Q}$ represents the text-formed \textit{question} created using multimodal experts, and $\mathcal{A}$ is the corresponding ground-truth AD, serving as the \textit{answer}. 
Then, for each test query $q$, we randomly sample $C$ demonstrations from $\mathcal{P}$ to facilitate the ICL process.


Furthermore, we are further interested in two essential questions: \textit{``What makes good examples for AD generation?''} and \textit{``How to find and use them for ICL?''} 


\Paragraph{Similarity-based MM-ICL.} A common approach, as suggested in existing literature~\cite{liu2021makes}, is to identify ``good examples'' based on similarity, employing a $k$-NN algorithm to select examples that exhibit the highest $k$ similarity between the embeddings of $\mathcal{Q}_i$ and the test query $q$. 
This solution expects to find supportive examples to benefit few-shot performance via a ``soft-copy'' mechanism~\cite{olsson2022context,han2023understanding}, which is often used in text classification tasks, such as sentiment analysis, or relatively-simple text generation task such as open-domain QA~\cite{dong2022survey}.

However, we empirically find that this similarity-based approach does not manage to enhance the ICL capability for AD generation, regardless of whether the retrieved examples are presented in descending order~\cite{liu2021makes} or ascending order~\cite{dong2022survey}.
We hypothesize that for complex text generation tasks such as AD generation, which requires multimodal perception and reasoning, similarity or relevance may not be the most suitable criteria for identifying effective ICL examples for improving overall performance. 

\begin{figure*}[th] 
    \centering
    \includegraphics[width=\linewidth]{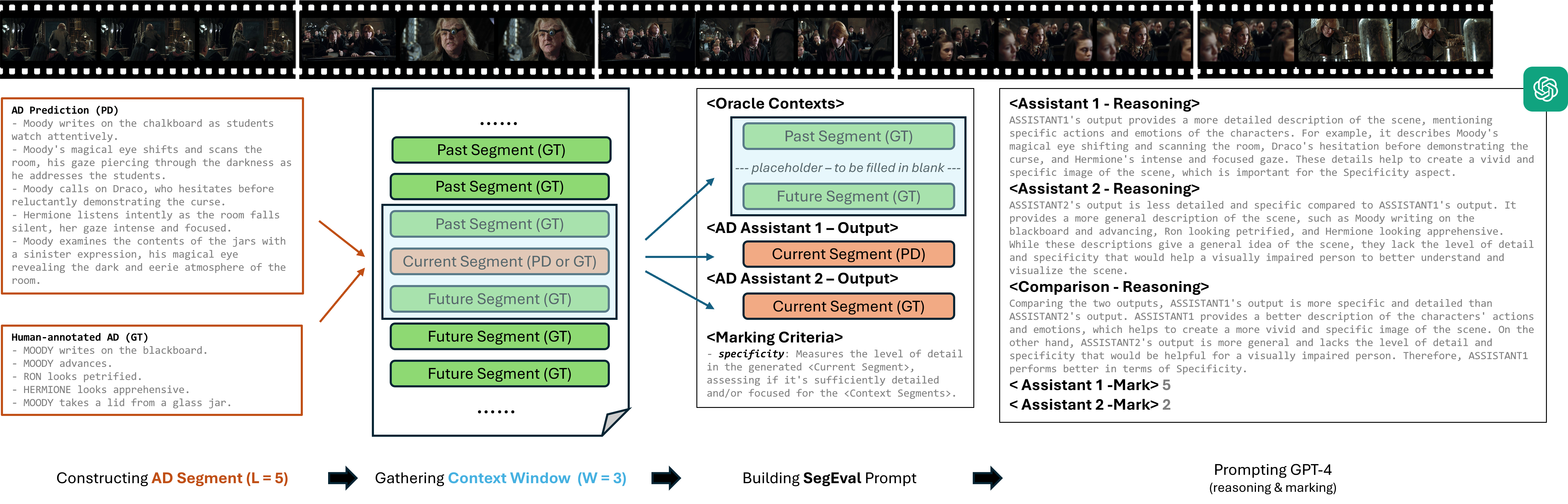}
    \caption{Our proposed $\mathtt{SegEval}$ evaluator to measure recurrent text generation quality with GPT-4 under customized marking criteria. Noticeably, GPT-4 is agnostic to the source of each assistant output (i.e., which $\mathtt{Seg}$ is GT or PD), and it would measure $\mathtt{Seg}$ quality taking oracle contexts into consideration. Take the response shown above as example, its corresponding re-scaled $r$ is 2.25. Zoom in for details.}
    \label{fig:segeval}
    \vspace{-0.35cm}
\end{figure*}

\Paragraph{Complexity-based MM-ICL.} Our empirical analysis reveals that not all questions are equally challenging, in terms of the complexity of multimodal fusion. Take Figure~\ref{fig:teaser} as an example: when comparing \textit{Titanic (1997)} to \textit{Spider Man (2018)}, the latter presents a more complex case. It requires the inference that ``Peter and Spider Man are the same character'', a deduction drawn from context AD and subtitles, alongside describing his actions from visual frames, enriched by contextual understanding from the context AD. 

This observation led us to hypothesize that complexity could be a more suitable metric for identifying effective ICL examples for tasks involving intricate multimodal fusion. To this end, we propose to query LLM to articulate the chain-of-thoughts (CoTs) as reasoning steps, denoted as $\mathcal{R}$, that assist in deriving the answer $\mathcal{A}$ from the question $\mathcal{Q}$. 
This process evolves our demonstration format from simple $(\mathcal{Q}, \mathcal{A})$ pairs to more comprehensive $(\mathcal{Q}, \mathcal{R}, \mathcal{A})$ tuples. 

Instead of the conventional random sampling from the entire pool $\mathcal{P}$, we propose selecting the most straightforward examples, quantified by the shortest number of reasoning steps. These are compiled into a simpler subset pool $\mathcal{P}_{simple}$, from which we conduct our demonstration sampling. This method ensures the inclusion of more intuitive and concise examples in our MM-ICL process. We present detailed ablation study in \S~\ref{subsec:abl_mmicl}, validating that complexity serves as a robust measure for selecting effective ICL examples for improving AD generation. 
\\
\\

\vspace{-10mm}
\section{Segment-based GPT-4 Evaluator}

The lack of standard AD annotation guidelines, varying cultural background and preferences of human annotators imply that AD is an inherently subjective recurrent text generation process, leading to notable inter-annotator disagreements~\cite{han2023autoad2} and challenges in evaluation using traditional reference-based captioning metrics.
To this end, inspired by~\cite{liu2023llava, liu2023gpteval}, we propose a segment-based GPT-4 evaluator $\mathtt{SegEval}$ to measure the recurrent AD generation, in terms of multi-domain qualities.

Suppose $L$ ADs form one segment $\mathtt{Seg}$.
For each $\mathtt{Seg}$, the evaluator takes into consideration an oracle context window $\mathtt{Ctx}$ of length $W$, to measure its multi-aspect scores. 
Specifically, we gather $W$ - 1 adjacent segments to form $\mathtt{Ctx}$, which consists of $\frac{W-1}{2}$ past and $\frac{W-1}{2}$ future segments surrounding the targeted $\mathtt{Seg}$.
Given a pair of predicted (PD) and ground-truth (GT) AD segments, $\mathtt{SegEval}$ would treat them as outputs of two separate AD generation systems, and query GPT-4 to reason and mark their raw marks independently.
The final score is calculated as the ratio $r$ of these raw marks between predicted and human-annotated AD, via post-processing. If the re-scaled $r$ is higher than 1.0, it indicates that GPT-4 might favour the predicted AD over human annotations under the specific aspect.  
Besides, this rescaling operation makes it comparable among different approaches, sharing human annotations as the marking standard.
Noticeably, although GPT-4 is unaware of the segment source that which $\mathtt{Seg}$ is the GT or PD, we always form $\mathtt{Ctx}$ from GT annotations to set the oracle for investigating contextual influences.

Overall, as shown in Figure~\ref{fig:segeval}, $\mathtt{SegEval}$ can measure context-irrelevant, short-context and long-context scores by flexibly changing the value of $W$. For example, it could measure \textit{text-level qualities} such as originality and consistency (when $W=1$), while it could also mark \textit{sequence-level qualities} such as coherence, diversity and specificity (when $W>1$). The details of each marking criteria are provided in appendix (\S~\ref{supp_sec:ad_eval}).

\newcommand{\myfulltab}{
  \tablestyle{0.6pt}{0.9} 
  \begin{tabular}{l l c c c c}
    \toprule[1.5pt]
       Method & Setting &  R-L ($\uparrow$) & C ($\uparrow$) & S ($\uparrow$) & R@5/16 ($\uparrow$)\\
       \hlineB{2.5}
       \multicolumn{5}{l}{\textit{Comparisons with fine-tuning-based SOTAs}} \\
        ClipCap~\cite{mokady2021clipcap} & Pretrain-Finetune  & 8.5  & 4.4 & 1.1 & 36.5\\
        ClipDec~\cite{nukrai2022text} & Pretrain-Finetune     & 8.2  & 6.7 & 1.4 & - \\
        AutoAD-I~\cite{han2023autoad1} & Pretrain-Finetune    & 11.9  & \textbf{14.3} & 4.4 & 42.1\\
        \textbf{\OURS{}}& Few-shot \& GPT-4  & \textbf{12.1} & 11.6 & \textbf{4.5} &\textbf{48.0} \\

        \textbf{\OURS{}}& Few-shot \& GPT-4V  & 10.1 & 4.9 & - & - \\

        \midrule
        \multicolumn{5}{l}{\textit{Comparisons with LLM/LMM baselines}} \\
        VLog\cite{VLog} & Zero-shot \& GPT-4 & 7.5 & 1.3 & 2.1 & 42.3\\
        MM-Vid~\cite{2023mmvid} & Zero-shot \& GPT-4V & 9.8 & 6.1 & 3.8 & 46.1\\
        \textbf{\OURSZS{}} & Zero-shot \& GPT-4 & 10.3 & 4.9  & 3.8 & 47.1 \\
        
        \midrule
        \multicolumn{5}{l}{\textit{Utilizing external character bank}} \\
        AutoAD-II \textcolor{gray}{\textdagger}~\cite{han2023autoad2} & \textcolor{gray}{Pretrain-Finetune}    & \textcolor{gray}{13.4}  & \textcolor{gray}{19.5} & \textcolor{gray}{-} & \textcolor{gray}{50.8}\\

        \textbf{\OURS{} \textcolor{gray}{\textdagger}}& \textcolor{gray}{Few-shot \& GPT-4} & \textcolor{gray}{13.4} & \textcolor{gray}{13.9} & \textcolor{gray}{5.2} &\textcolor{gray}{49.0}\\
        
        \textbf{\OURS{} \textcolor{gray}{\textdagger}}& \textcolor{gray}{Few-shot \& GPT-4V} & \textcolor{gray}{12.3} & \textcolor{gray}{8.3} & \textcolor{gray}{-} &\textcolor{gray}{-}\\

    \bottomrule[1.5pt]
    \end{tabular}
}

\section{Experiments}
\label{sec:results}

\subsection{Evaluation Setup}
\Paragraph{Datasets.} 
We conduct experiments on the AD generation benchmark established in AutoAD~\cite{han2023autoad1}, where \MADTRAIN{} and \MADEVAL{} are released as training and testing splits, respectively.
\textbf{\MADTRAIN{}} consists of 334,296 ADs and 628,613 subtitles from 488 movies, while \textbf{\MADEVAL{}} is compromised of 6,520 ADs and 10,602 subtitles from 10 movies.

\Paragraph{Metrics.}
Following AutoAD\cite{han2023autoad1}, we report three traditional captioning metrics to measure the quality of ADs generated versus human-annotated ones, including ROUGE-L~\cite{Lin_2004} (\textbf{R-L}), CIDEr~\cite{Vedantam_Zitnick_Parikh_2015} (\textbf{C}) and SPICE~\cite{Anderson_Fernando_Johnson_Gould_2016} (\textbf{S}).
Besides, we follow AutoAD-II to benchmark the text sequence generation over their recall-based metric `Recall@k within Neighbours' (\textbf{R@k/N}), where the text similarity is measured by BertScore~\cite{zhang2019bertscore}.
We also report Bleu-1~\cite{papineni2002bleu} and METEOR~\cite{banerjee2005meteor} for ablation studies.
Each experiment of \OURS{} is repeated \textit{three} times in ~\Cref{tab:sota_comp,tab:sota_comp_llm,tab:sota_comp_external,tab:mmicl,tab:ours_with_vision,tab:GPT-4eval}, as well as Figure~\ref{fig:improvement}, with $\texttt{mean}$ (and $\texttt{std}$) reported.

\subsection{Comparison with State-of-the-Art Approaches}




\Paragraph{Fine-tuning-based SOTAs.}
 We first compare our training-free framework against the fine-tuning-based SOTAs, including ClipCap~\cite{mokady2021clipcap}, ClipDec~\cite{nukrai2022text} and AutoAD-I~\cite{han2023autoad1}.
As shown in Table~\ref{tab:sota_comp}, our training-free approach outperforms its fine-tuning-based counterparts~\cite{mokady2021clipcap, nukrai2022text, han2023autoad1}, in terms of ROUGE-L, SPICE and R@k/N, especially the AutoAD-I~\cite{han2023autoad1} (R-L 12.1 vs 11.9; S 4.5 vs 4.4; R@k/N 48.0 vs 42.1) which is proposed to conduct partial data pretraining over an extra large-scale text-only AV-AD dataset~\cite{AVAD,han2023autoad1} (consisting of 3.3M ADs from over 7k movies) to address the lack of paired training data for AD generation.
Unlike ~\cite{han2023autoad1, han2023autoad2} who report to struggle with benefiting from character dialogues, our \OURS{} could better integrate multimodal information and effectively identify characters from appropriate subtitle usage (shown as \S~\ref{subsec:improvement}-D).



\Paragraph{LLM/LMM Baselines.}
We next compare our \OURS{} with LLM and LMM baselines:
(a) VLog~\cite{VLog} is a LLM-based method for long video understanding. It converts multimodal perceptions into natural languages via several pretrained models (BLIP-2~\cite{li2023blip2}, GRIT~\cite{wu2022grit} and Whisper~\cite{radford2023robust}), and then utilizes a LLM to generate texts based on task-specific prompts. To make a fair comparison, we make it query GPT-4 (rather than GPT-3.5)  with the same AD generation prompt as we use in \OURS{}.
(b) MM-Vid~\cite{2023mmvid} is a LMM system which generates AD through incorporating external knowledge with clip-level video description generated by GPT-4V~\cite{gpt4v,yang2023dawn}. 

As shown in Table~\ref{tab:sota_comp_llm}, our \OURS{} (\textit{w/o MM-ICL}) would outperform VLog, which is mainly attributed to the proposed short-term memory queue and long-term visual memory to effectively leverage relevant contextual information recalled from past ADs. 
In addition, while \OURS{} is based on GPT-4 (text-only), it  
also surpasses the GPT-4V(ision) based MM-Vid system in terms of R-L and SPICE. 
The results suggest that a memory-augmented LLM can be comparably valuable to the perception-enhanced ones. 
Furthermore, with our proposed \MMICL, \OURS{} outperforms VLog and MM-Vid by a large margin.

\Paragraph{Utilizing External Character Bank.}
Previously, all discussed methods share the same and only knowledge source to assist in character recognition. 
More specifically, they, like us humans, mostly identify characters and infer their names through hearing (i.e., auditory cues) \textit{alone} when watching movies. 
Given this single source of gaining character information, our \OURS{} would convey contextual information via retrieving visual and temporal memories. 
However, these methods suffer from an unavoidable limitation: The character identities would unfortunately remain mystery until their names are being first-time called in dialogues. 

To alleviate that, following AutoAD-II~\cite{han2023autoad2} we also investigate how our method could benefit from incorporating an external character bank.
To construct this character bank, ~\cite{han2023autoad2} exploits actor portrait images (from an external movie database) to retrieve a few most similar frames for each main character in each movie.
Unlike ~\cite{han2023autoad2} who trains an auxiliary character recognizer from these retrieved frames, we maintain our training-free designs by simply concatenating these frames into short video clips to introduce each character (with ADs as their names). 
Next, we prepend these video clips to the long-form videos, such that they could work compatibly with our register-and-recall mechanism.
As shown in Table~\ref{tab:sota_comp_external}, our \OURS{} (\textit{w/ ExtCharBank}) could further boost its performance and generate outcomes comparable to the fine-tuning-based AutoAD-II.

\begin{table}
    \centering
\tablestyle{2.5pt}{1.0} 
  \begin{tabular}{l c c c c c}
    \toprule[1.5pt]
       Method & Training-Free &  R-L ($\uparrow$) & C ($\uparrow$) & S ($\uparrow$) & R@5/16 ($\uparrow$)\\
       \midrule
        ClipCap~\cite{mokady2021clipcap} & \ngmark  & 8.5  & 4.4 & 1.1 & 36.5\\
        ClipDec~\cite{nukrai2022text} & \ngmark     & 8.2  & 6.7 & 1.4 & - \\
        AutoAD-I~\cite{han2023autoad1} & \ngmark    & 11.9  & \textbf{14.3} & 4.4 & 42.1\\
        \midrule
        \textbf{\OURS{}}& \okmark  & \textbf{12.1} & 11.6 & \textbf{4.5} &\textbf{48.0} \\

    \bottomrule[1.5pt]
    \end{tabular}

    \caption{Comparisons with fine-tuning-based state-of-the-art methods on \MADEVAL{} benchmark. Note: the random guess will result in a R@5/16 of 31.3\%.
    }
    \vspace{-0.25cm}
    \label{tab:sota_comp}
\end{table}

\begin{table}
    \centering
\tablestyle{3.5pt}{1} 
  \begin{tabular}{l l c c c c}
    \toprule[1.5pt]
       Method & LLM &  R-L ($\uparrow$) & C ($\uparrow$) & S ($\uparrow$) & R@5/16 ($\uparrow$)\\
       \midrule
        VLog\cite{VLog} & GPT-4 & 7.5 & 1.3 & 2.1 & 42.3\\
        MM-Vid~\cite{2023mmvid} & GPT-4V & 9.8 & 6.1 & 3.8 & 46.1\\
        \midrule
        \textbf{\OURS{}} \\ 
         \ \ \textit{w/o} MM-ICL & GPT-4 & 10.3 & 4.9  & 3.8 & 47.1 \\
        \ \ \textit{w/} MM-ICL &  GPT-4  & \textbf{12.1} & \textbf{11.6} & \textbf{4.5} & \textbf{48.0} \\

    \bottomrule[1.5pt]
\end{tabular}
    \caption{Comparisons with training-free LLM/LMM baselines on \MADEVAL{} benchmark.
    }
    \vspace{-0.25cm}
    \label{tab:sota_comp_llm}
\end{table}

\begin{table}
    \centering
\tablestyle{2.6pt}{1.0} 
  \begin{tabular}{l c c c c c}
    \toprule[1.5pt]
       Method & Training-Free &  R-L ($\uparrow$) & C ($\uparrow$) & S ($\uparrow$) & R@5/16 ($\uparrow$)\\
       \midrule
        AutoAD-II \textdagger~\cite{han2023autoad2} & \ngmark    & 13.4  & 19.5 & - & 50.8\\
        \textbf{\OURS{} \textdagger}& \okmark & 13.4 & 13.9 & 5.2 &49.0\\
        
    \bottomrule[1.5pt]
\end{tabular}
    \caption{Evaluation on \MADEVAL{} benchmark, with an external character bank annotated and utilized for improved character recognition (denoted as \textdagger{}).
    }
    \vspace{-0.25cm}
    \label{tab:sota_comp_external}
\end{table}

\begin{figure*}[t] 
    \centering
    \includegraphics[width=\linewidth]{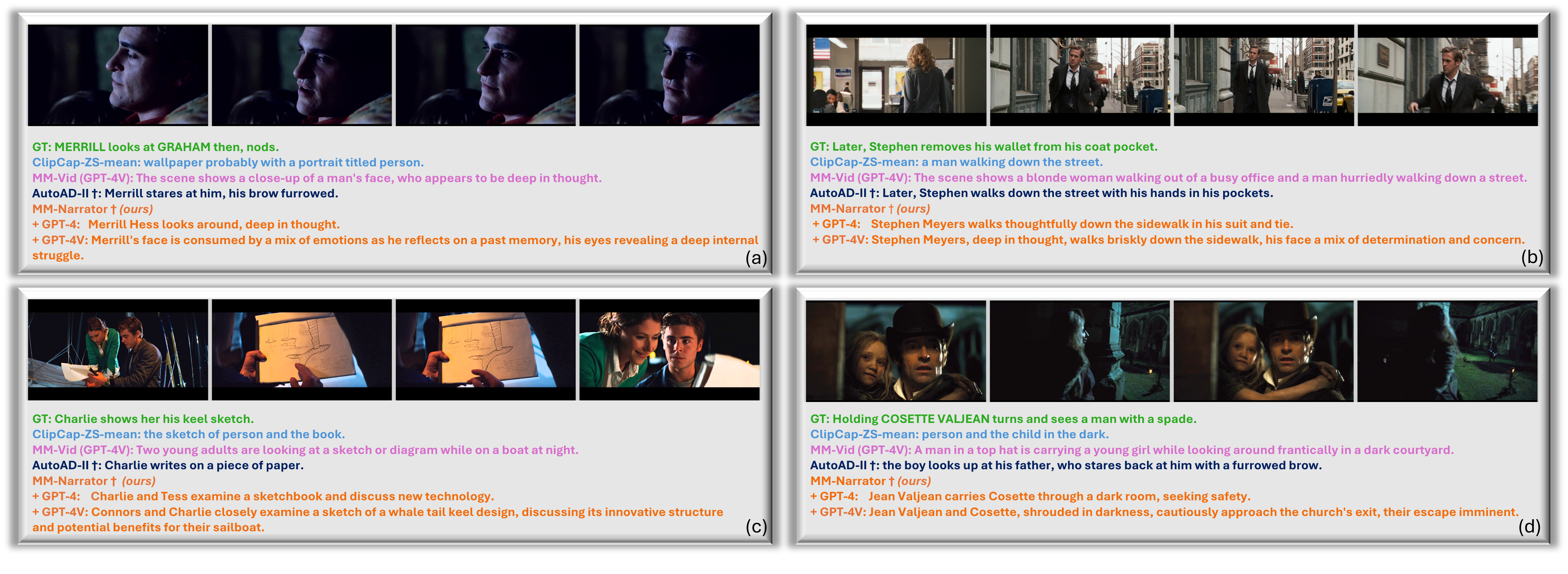}
    \caption{Qualitative comparisons between ClipCap, MM-Vid, AutoAD-II, and our \OURS{}, where the latter two approaches are equipped with the external character bank. The movies are from (a) \textit{Signs (2002)}, (b) \textit{Ides of March (2011)}, (c) \textit{Charlie St. Cloud (2010)}, and (d) \textit{Les Misérables (2012)}. Zoom in for details.}
    \label{fig:qual_analysis}
    \vspace{-0.25cm}
\end{figure*}

\Paragraph{Qualitative Results.} Qualitative comparisons over MAD-eval dataset are shown as Figure~\ref{fig:qual_analysis}, while the qualitative demonstrations of applying our \OURS{} on other long-form videos (external to the MAD-eval dataset) are shown in Figure~\ref{fig:teaser}.
Additional qualitative comparisons and demonstrations are included in the appendix (Figure~\ref{fig:more_qual_dataset} and~\ref{fig:more_qual_extra}).

\subsection{Building \OURS{} From Image Captioner}
\label{subsec:improvement}
\begin{figure}[t] 
    \centering
    \includegraphics[width=\linewidth]{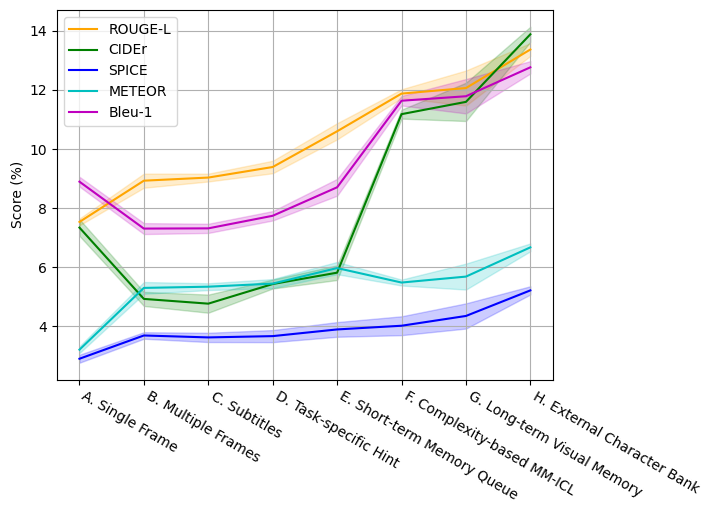}
    \caption{Ablations on each component for \OURS{}.}
    \label{fig:improvement}
    \vspace{-0.25cm}
\end{figure}

As shown in Figure~\ref{fig:improvement}, we quantitatively demonstrate how our training-free \OURS{} are developed step by step.
Starting from (A) an image captioner, we elaborate how multimodal perception benefits \OURS{} to form an intricate multimodal understanding over video content. 
Specifically, it includes adding (B) multiple frames, (C) subtitles, and (D) a task-specific hint\footnote{\textit{``Hint: try to infer character names from subtitles for AD generation."}}.
Noticeably, simply adding the dialogues (C) might not result in an immediate performance gain. 
However, with prompt engineering in (D), \OURS{} pays more attention to effectively leverage multimodal clues for character-centric AD generation.

Next, we illustrate how we transform \OURS{} into recurrent AD narrator to produce story-coherent AD, with incorporation of past memories and complexity-based MM-ICL.
Specifically, \OURS{} maintains (E) a short-term memory queue, learns from (F) multimodal demonstrations via MM-ICL, and retrieves (G) long-term visual memory for character re-identification, which could be further boosted with (H) an external character bank.

\begin{table}
    \centering
    \tablestyle{1.0pt}{1.0} 
    \begin{tabular}{l l l c c c c }
    \toprule[1.5pt]
       \multirow{2}{*}{Model}  & \multirow{2}{*}{Pool Size} & Demo. & \multirow{2}{*}{CoT} & \multirow{2}{*}{R-L ($\uparrow$)} &\multirow{2}{*}{C ($\uparrow$)} & \multirow{2}{*}{B-1 ($\uparrow$)}\\
       &  & Format & & \\
       \hlineB{2.5}
       \multicolumn{5}{l}{\textit{\textbf{R}andom MM-ICL}} \\
       \rowcolor{lightgray} R1~\cite{brown2020language} & 100\% & $(\mathcal{Q, A})$ & \ngmark & 13.2$_{\pm0.1}$ & 12.9$_{\pm0.2}$ & 12.2$_{\pm0.1}$\\
       R2 & 100\% & $(\mathcal{Q, R, A})$ & \okmark & 13.4$_{\pm0.1}$ &13.4$_{\pm0.2}$ &12.7$_{\pm0.1}$ \\
       R3 & 10\% \textit{random} & $(\mathcal{Q, A})$ & \ngmark &  13.3$_{\pm0.1}$ &13.0$_{\pm0.1}$ & 12.3$_{\pm0.0}$\\
       R4 & 10\% \textit{random} & $(\mathcal{Q, R, A})$ & \okmark &  13.3$_{\pm0.1}$ &13.4$_{\pm0.1}$ & 12.6$_{\pm0.0}$\\
        \midrule
        \multicolumn{5}{l}{\textit{\textbf{S}imilarity-based MM-ICL}} \\
        \rowcolor{lightgray}S1~\cite{liu2021makes} & 100\% & $(\mathcal{Q, A})$ & \ngmark & 13.5$_{\pm0.0}$ &13.1$_{\pm0.0}$ & 12.6$_{\pm0.1}$\\

        \midrule
        \multicolumn{5}{l}{\textit{\textbf{C}omplexity-based MM-ICL}} \\
        C1 & 10\% \textit{shortest} & $(\mathcal{Q, A})$ & \ngmark & 13.2$_{\pm0.1}$ &13.3$_{\pm0.3}$ & 12.3$_{\pm0.1}$\\
        \rowcolor{lightgray}C2 \textbf{\textit{(ours)}} & 10\% \textit{shortest} &  $(\mathcal{Q, R, A})$ & \okmark & 13.4$_{\pm0.0}$ &13.9$_{\pm0.1}$ & 12.8$_{\pm0.0}$\\
        C3 & 10\% \textit{longest}  & $(\mathcal{Q, R, A})$ & \okmark & 13.3$_{\pm0.1}$ &12.7$_{\pm0.2}$ & 12.4$_{\pm0.1}$\\ 

    \bottomrule[1.5pt]
    \end{tabular}
    \caption{Different MM-ICL designs for \OURS{}~\textdagger, with each representative implementation highlighted.}
    \label{tab:mmicl}
    \vspace{-0.25cm}
\end{table}

\subsection{Ablations on Multimodal In-Context Learning}
\label{subsec:abl_mmicl}

\begin{table*}[th]
    \centering
    \scalebox{0.87}{
    \begin{tabular}{|l | l | l l | l l l | l l l|  }
           \hline

       \multirow{4}{*}{Method}  & \multirow{4}{*}{LLM/LMM} &\multicolumn{2}{c|}{\textbf{Text-level Quality}} & \multicolumn{6}{c|}{\textbf{Sequence-level Quality}} \\
        & &\multicolumn{2}{c|}{Context-irrelevant Scores} & \multicolumn{3}{c}{Short-context Scores} & \multicolumn{3}{c|}{Long-context Scores}\\
        \cline{3-10}
        &  & $\mathtt{Orig.}$ & $\mathtt{Cons.}$ & $\mathtt{Cohe.}$ & $\mathtt{Dive.}$ & $\mathtt{Spec.}$ & $\mathtt{Cohe.}$ & $\mathtt{Dive.}$ & $\mathtt{Spec.}$ \\
        &  & \textcolor{gray}{$_{\pm{0.02}}$} & \textcolor{gray}{$_{\pm{0.02}}$} & \textcolor{gray}{$_{\pm{0.01}}$} & \textcolor{gray}{$_{\pm{0.06}}$} & \textcolor{gray}{$_{\pm{0.04}}$} &\textcolor{gray}{$_{\pm{0.01}}$} & \textcolor{gray}{$_{\pm{0.01}}$} & \textcolor{gray}{$_{\pm{0.03}}$} \\

       \hline
       \textcolor{gray}{GT} & \textcolor{gray}{-} & \textcolor{gray}{1.00} & \textcolor{gray}{1.00} & \textcolor{gray}{1.00} & \textcolor{gray}{1.00} & \textcolor{gray}{1.00} & \textcolor{gray}{1.00} & \textcolor{gray}{1.00} & \textcolor{gray}{1.00} \\
        ClipCap~\cite{mokady2021clipcap} & GPT-2 & 0.43 & 0.42 & 0.26 & 0.35 & 0.35 & 0.26 & 0.42 & 0.33\\
        VLog\cite{VLog}  & GPT-4 & 1.03 & 0.88 & 0.34  & 0.55 & 0.52 & 0.32 & 0.57 & 0.43\\
        MM-Vid\cite{2023mmvid}  & GPT-4V & 0.85 & 0.78 & 0.51  & 0.81 & 0.66 & 0.53 & 0.84 & 0.62\\

        \hline
        \textbf{\OURS{}}  & GPT-4 & 1.05$_{\pm0.10}$ & 1.03$_{\pm0.05}$ & 0.52$_{\pm0.06}$ & 0.70$_{\pm0.06}$ & 0.66$_{\pm0.04}$ & 0.57$_{\pm0.05}$ & 0.70$_{\pm0.02}$& 0.61$_{\pm0.05}$\\
        \textbf{\OURS{}}  & GPT-4V & \textbf{1.49$_{\pm0.10}$} & 1.45$_{\pm0.05}$ & 
        0.94$_{\pm0.07}$ & 1.01$_{\pm0.04}$  & 1.13$_{\pm0.08}$
        & 0.87$_{\pm0.04}$ & 1.05$_{\pm0.04}$
         &  \textbf{1.14$_{\pm0.05}$}\\
        \textbf{\OURS{} \textdagger} & GPT-4 & 0.95$_{\pm0.02}$ & 1.06$_{\pm0.01}$ & 0.62$_{\pm0.04}$ & 0.75$_{\pm0.01}$ & 0.76$_{\pm0.01}$ & 0.62$_{\pm0.04}$ & 0.80$_{\pm0.03}$ & 0.71$_{\pm0.03}$\\

        \textbf{\OURS{} \textdagger} & GPT-4V & {1.45$_{\pm0.14}$} & \textbf{1.46$_{\pm0.04}$} & \textbf{0.98$_{\pm0.03}$} & \textbf{1.06$_{\pm0.04}$} & \textbf{1.24$_{\pm0.09}$} & \textbf{0.94$_{\pm0.02}$} &\textbf{1.09$_{\pm0.05}$} & 1.12$_{\pm0.03}$\\
           \hline

    \end{tabular}
    }
    \caption{Evaluating AD generation with $\mathtt{SegEval}$ on \MADEVAL{} benchmark, with segment size $L$ set to 5. The context window sizes $W$ are set as 1 / 3 / 11 to compute context-irrelevant / short-context / long-context scores, respectively. $\mathtt{Orig.}$, $\mathtt{Cons.}$, $\mathtt{Cohe.}$, $\mathtt{Dive.}$, and $\mathtt{Spec.}$ stand for \textit{originality}, \textit{consistency}, \textit{coherence}, \textit{diversity}, and \textit{specificity}, respectively. The scoring variances of these GPT-4 evaluators are denoted below for references, which are estimated by three repeated evaluations over the same inference outputs. These re-scaled scores measure the corresponding AD prediction (PD) qualities of each specific method, compared to the shared marking standards set by ground-truth (GT) ADs. For example, given a pair of PD and GT segments, without revealing to the evaluator which segment is GT or PD, if it reasons and marks the raw qualities ($\mathtt{R.Q.}$) as 8 and 5 for PD and GT segments, respectively, we derive the re-scaled score $r$ as $\frac{\mathtt{R.Q.}_{\mathtt{PD}}}{\mathtt{R.Q.}_{\mathtt{GT}}} = \frac{8}{5} = 1.6$. \textdagger{} indicates our incorporation with ExtCharBank.} 
    \label{tab:GPT-4eval}
    \vspace{-0.25cm}
\end{table*}

We investigated three groups of MM-ICL, including random, similarity-based, and our proposed complexity-based MM-ICL. 
The results as shown in Table~\ref{tab:mmicl}, verify our hypothesis that complexity serves as an appropriate measure for selecting effective ICL demonstrations for improving AD generation.
Below, we further discuss three sub-questions to elaborate an in-depth analysis.

\Paragraph{Does CoT help?}
We propose to adopt CoT technique to obtain the intermediate reasoning steps $\mathcal{R}$ that help derive answer $\mathcal{A}$ from question $\mathcal{Q}$. 
This automatic process extends demonstration format from $(\mathcal{Q}, \mathcal{A})$ pairs to $(\mathcal{Q}, \mathcal{R}, \mathcal{A})$ tuples.
As its consistent gains could be observed multiple times (R1 vs R2; R3 vs R4; C1 vs C2), adding multimodal multi-step reasoning $\mathcal{R}$ during MM-ICL could help \OURS{} improve its multimodal reasoning capability to better incorporate multimodal inputs.
Qualitative demonstrations of $\mathcal{R}$ are shown as Figure~\ref{fig:CoT} in appendix.

\Paragraph{Does complexity-based ranking help?}
We observed that conducting MM-ICL with the most intuitive examples benefits the overall performance (R4 vs C2), however, switching with the hardest ones which own the longest reasoning steps, MM-ICL actually leads to a decline in performance (R4 vs C3).
These results indicate that more straightforward examples, quantified by the shortest number of reasoning steps,
compile to a simpler yet more powerful subset MM-ICL demonstration pool for effective AD generation.

\Paragraph{Does complexity-based MM-ICL work effectively?}
Combining CoT with complexity-based ranking, our proposed complexity-based MM-ICL (C2) performs more favorably than the random and similarity-based sampling approaches (R1~\cite{brown2020language} and S1~\cite{liu2021makes}), which are classic solutions in choosing few-shot ICL examples.
Besides, ours is easy-to-implement and explainable-to-human, avoiding the computation overhead of retrieval-based selection.
More discussions can be found in appendix (\S~\ref{supp_sec:mm_icl}).


\subsection{Evaluating AD Generation with GPT-4}

\begin{table}
    \centering
\tablestyle{1.1pt}{0.9} 
\begin{tabular}{l c c c c}
    \toprule[1.5pt]
       Method  &  R-L ($\uparrow$) & C ($\uparrow$) & M ($\uparrow$) & B-1 ($\uparrow$)\\
       \midrule
        \textbf{\OURS{}} \\
        + GPT-4  & $12.1_{\pm0.4}$ & $11.6_{\pm0.4}$ & $5.7_{\pm0.2}$ &$11.8_{\pm0.3}$ \\
        + GPT-4V & $11.8_{\pm0.1}$ & $7.0_{\pm0.2}$ & $6.5_{\pm0.1}$ & $9.3_{\pm0.1}$ \\
        \midrule
        \textbf{\OURS{} \textdagger}\\
        + GPT-4& $13.4_{\pm0.0}$ &$13.9_{\pm0.1}$& $6.7_{\pm0.0}$ & $12.8_{\pm0.0}$\\
        + GPT-4V &$12.8_{\pm0.0}$ & $9.8_{\pm0.2}$& $ 7.1_{\pm0.0}$ &$10.9_{\pm0.0}$\\
    \bottomrule[1.5pt]
\end{tabular}
    \caption{Comparisons over classic reference-based captioning scores, when incorporating our \OURS{} with GPT-4V.}
    \label{tab:ours_with_vision}
    \vspace{-0.25cm}
\end{table}

In Table~\ref{tab:ours_with_vision}, we observe a few performance drop on classic reference-based captioning scores when incorporating \OURS{} with GPT-4V~\cite{gpt4v}. As shown in Figure~\ref{fig:qual_analysis}, the decrease in performance can be primarily attributed to the more detailed and much richer ADs generated by our method, which diverge from the typically shorter human-annotated ADs in \MADEVAL{}.
This suggests that taking human annotated AD as oracles to measure AD-level captioning scores might be unsuitable for advanced LMM approaches, which further motivates our proposal of evaluating recurrent text generation with GPT-4.

Adjusting $W$, our proposed $\mathtt{SegEval}$ could flexibly measure both \textit{text-level} and \textit{sequence-level} qualities.
As shown in Table~\ref{tab:GPT-4eval}, 
the performance ranking order observed in $\mathtt{SegEval}$ aligns with our other experimental results, validating the reliability of $\mathtt{SegEval}$  as an evaluation tool, except for GPT-4V based MM-Vid where ours falls short on \textit{diversity}.
Furthermore, when employing GPT-4V as our vision expert, \OURS{} not only outperforms others by a large margin, but also closely mirrors the quality of human annotated ADs in multiple aspects, gaining more favor from the source-agnostic GPT-4 evaluator.

Compared to classic reference-based captioning scores, $\mathtt{SegEval}$ could better reflect the recurrent text generation qualities with GPT-4. 
One human validation on $\mathtt{SegEval}$ is shown in Figure~\ref{fig:segeval}, and more examples can be found in appendix (Figure~\ref{fig:eval_breakdown}).
Moreover, $\mathtt{SegEval}$ could be easily extended to support more comprehensive evaluation perspectives by querying it with extra customized marking criteria.

\section{Conclusion}
\OURS{} represents a significant leap in automatic audio description (AD) generation for long-form videos, leveraging the power of GPT-4 and innovative multimodal in-context learning (MM-ICL).
This recurrent AD narrator excels in generating story-coherent and character-centric AD by combining immediate textual context with long-term visual memory.
Its training-free design, coupled with our proposed complexity-based MM-ICL demonstration selection strategy, outperforms both existing fine-tuning-based and LLM-based approaches in most scenarios, as measured by traditional captioning metrics.
Furthermore, we introduce a GPT-4 empowered evaluator for a more comprehensive measurement of recurrent text generation qualities.
Its results suggest that \OURS{} generates AD comparable to human annotations across several considered aspects.


{\small
\vspace{-1pt}
\subsection*{Acknowledgment}
\vspace{-2pt}

We are deeply grateful to OpenAI for providing access to their exceptional tool~\cite{gpt4,gpt4v}. We also extend heartfelt thanks to our Microsoft colleagues for their insights, with special acknowledgment to Faisal Ahmed, Ehsan Azarnasab, and Lin Liang for their constructive feedback.

\bibliographystyle{ieee_fullname}
\bibliography{./egbib.bib}
}

\clearpage 
\appendix
\section*{Appendix} 

In this appendix, we present more details and discussions of AD generation with \OURS{} (\S\ref{supp_sec:ad_gen}), our proposed complexity-based MM-ICL (\S\ref{supp_sec:mm_icl}) and AD evaluation with $\mathtt{SegEval}$ (\S\ref{supp_sec:ad_eval}).
Next, we elaborate our implementation details (\S\ref{supp_sec:impl}) and discuss the future work on both AD generation and evaluation (\S\ref{supp_sec:future}).

\section{AD Generation}
\label{supp_sec:ad_gen}
\begin{figure*}[th] 
    \centering
    \includegraphics[width=\linewidth]{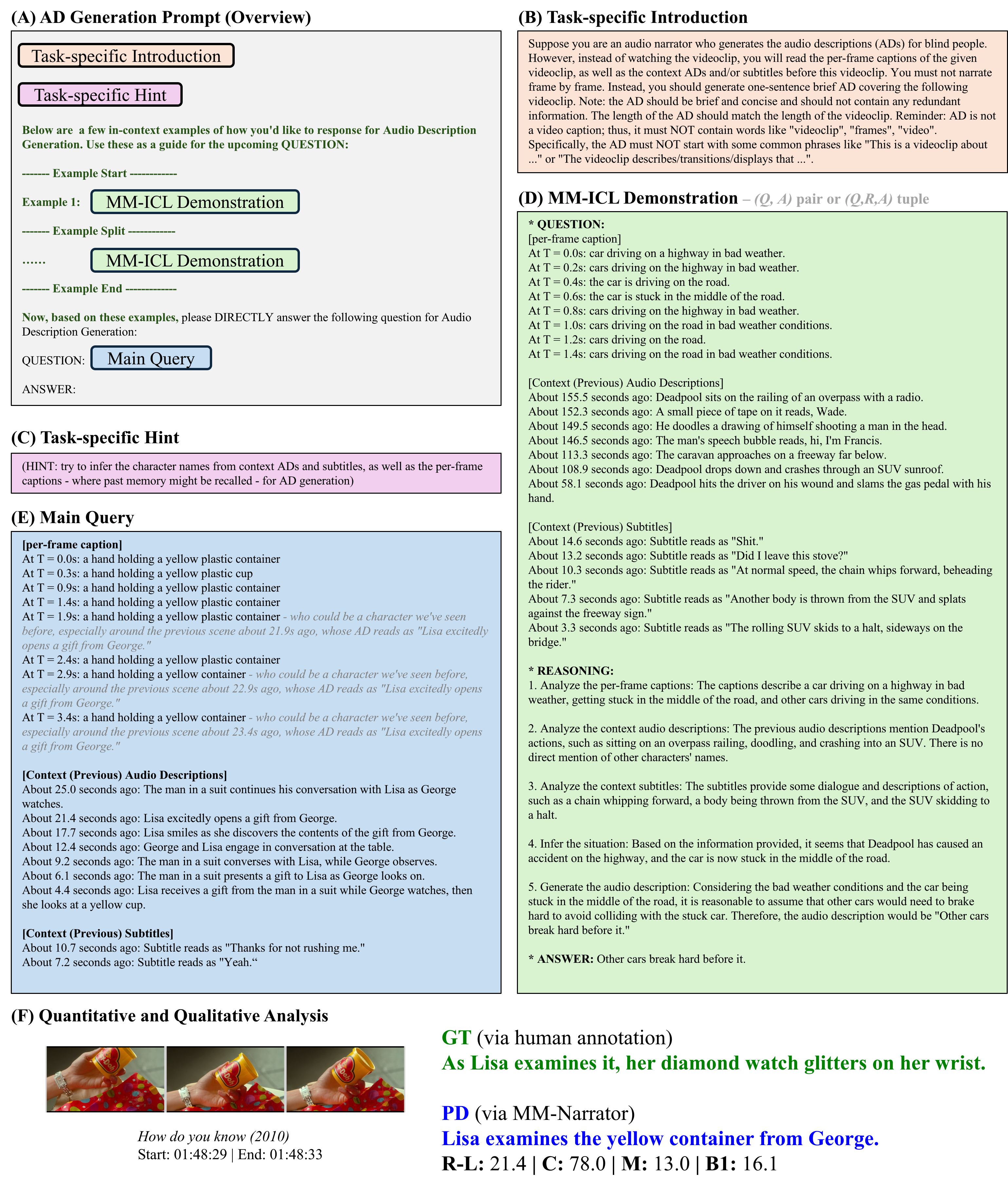}
    \caption{A breakdown of the AD generation prompt constructed by \OURS{}, including an (A) overview with ICL-specific instructions marked in green, (B) task-specific introduction $I_\mathtt{task}$ and (C) hint $H_\mathtt{task}$, a few multimodal ICL (MM-ICL) demonstrations $\mathcal{D}_\mathtt{ICL}$ with an example shown as (D), and (E) the main query $q_\mathtt{main}$ to be responded by GPT-4, with long-term visual memory marked in gray. Eventually, we show the corresponding (F) quantitative and qualitative analysis of the AD prediction via \OURS{} against the human AD annotation. Zoom in for details.}

    \label{fig:breakdown}
\end{figure*}

\OURS{} builds prompts to query GPT-4 for recurrent AD generation, including the following elements: task-specific introduction $I_\mathtt{task}$ and hint $H_\mathtt{task}$, main query $q_\mathtt{main}$, as well as a set of few-shot multimodal demonstrations $\mathcal{D}_\mathtt{ICL}$ to conduct in-context learning. 
With a breakdown shown in Figure~\ref{fig:breakdown}, we present the details as follows.

\Paragraph{Querying with multimodal clues.}
Both the main query $q_\mathtt{main}$ and the demonstration queries in $\mathcal{D}_\mathtt{ICL}$ are formatted with the same query builder, which outputs AD query from multiple text-formed multimodal clues. These multimodal clues include visual captions ($\boldsymbol{x}^{cap}_{i}$) with successfully re-identified characters, recent context ADs ($\mathcal{M}_{\boldsymbol{short}}$) and character dialogues ($\boldsymbol{x}^{sub}_{t\in T_{sub}}$).

\Paragraph{Prompting with MM-ICL.}
Each MM-ICL demonstration within $\mathcal{D}_\mathtt{ICL}$, is composed of a pair $(\mathcal{Q}, \mathcal{A})$ or a tuple $(\mathcal{Q}, \mathcal{R}, \mathcal{A})$ when chain-of-thought (CoT) is adopted to generate the multimodal multi-step reasoning $\mathcal{R}$ that derives answer $\mathcal{A}$ from question $\mathcal{Q}$.

\begin{figure*}[th] 
    \centering
    \includegraphics[width=\linewidth]{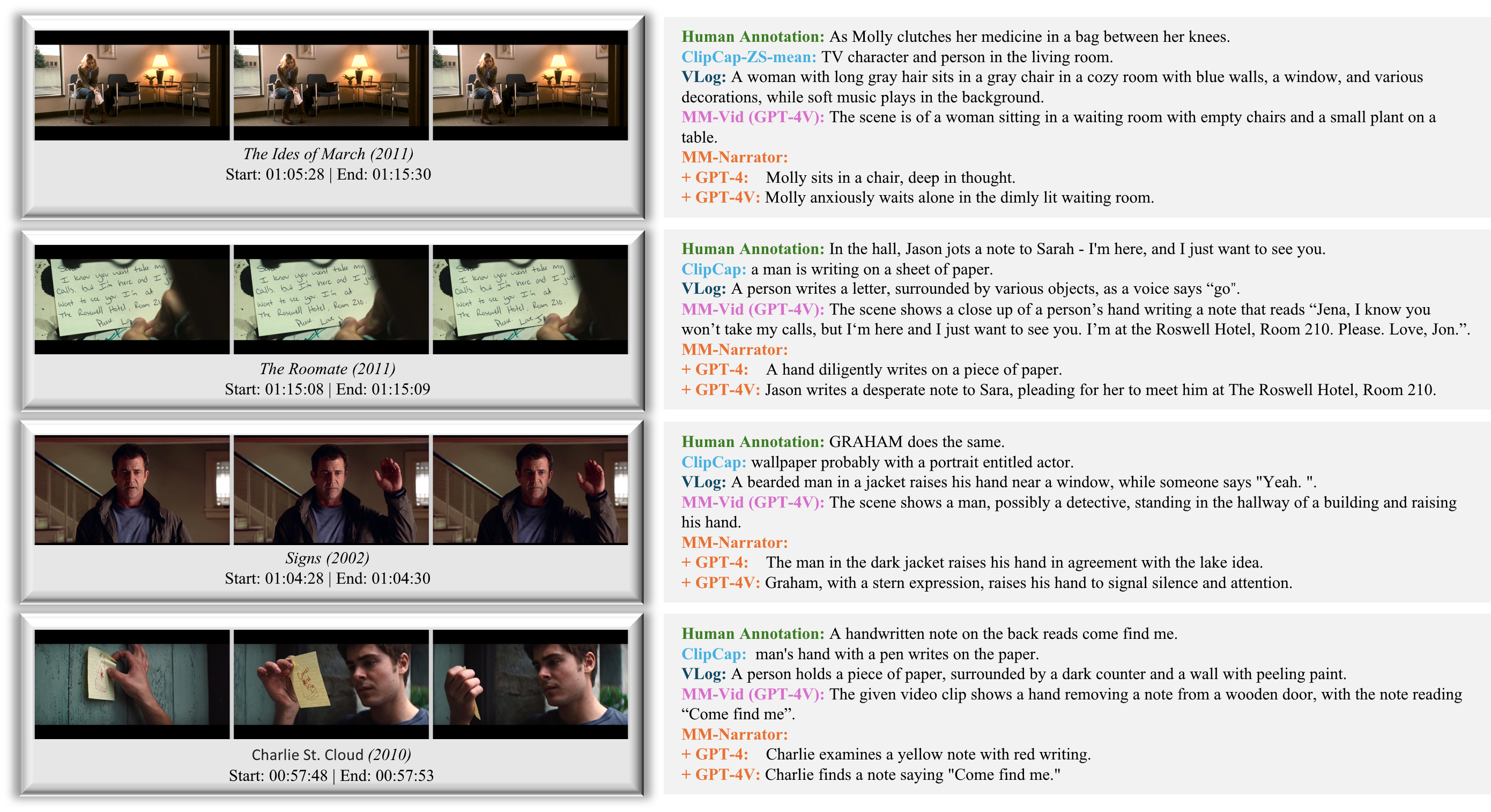}
    \caption{More qualitative comparisons on \MADEVAL{} benchmark. For example, in \textit{The ides of March (2011)}, our method generates AD by conditioning on current video clip and the contextual information from timestamp 00:00:00 to 01:05:28. Zoom in for details.} 
    \label{fig:more_qual_dataset}
\end{figure*}

\begin{figure*}[th] 
    \centering
    \includegraphics[width=\linewidth]{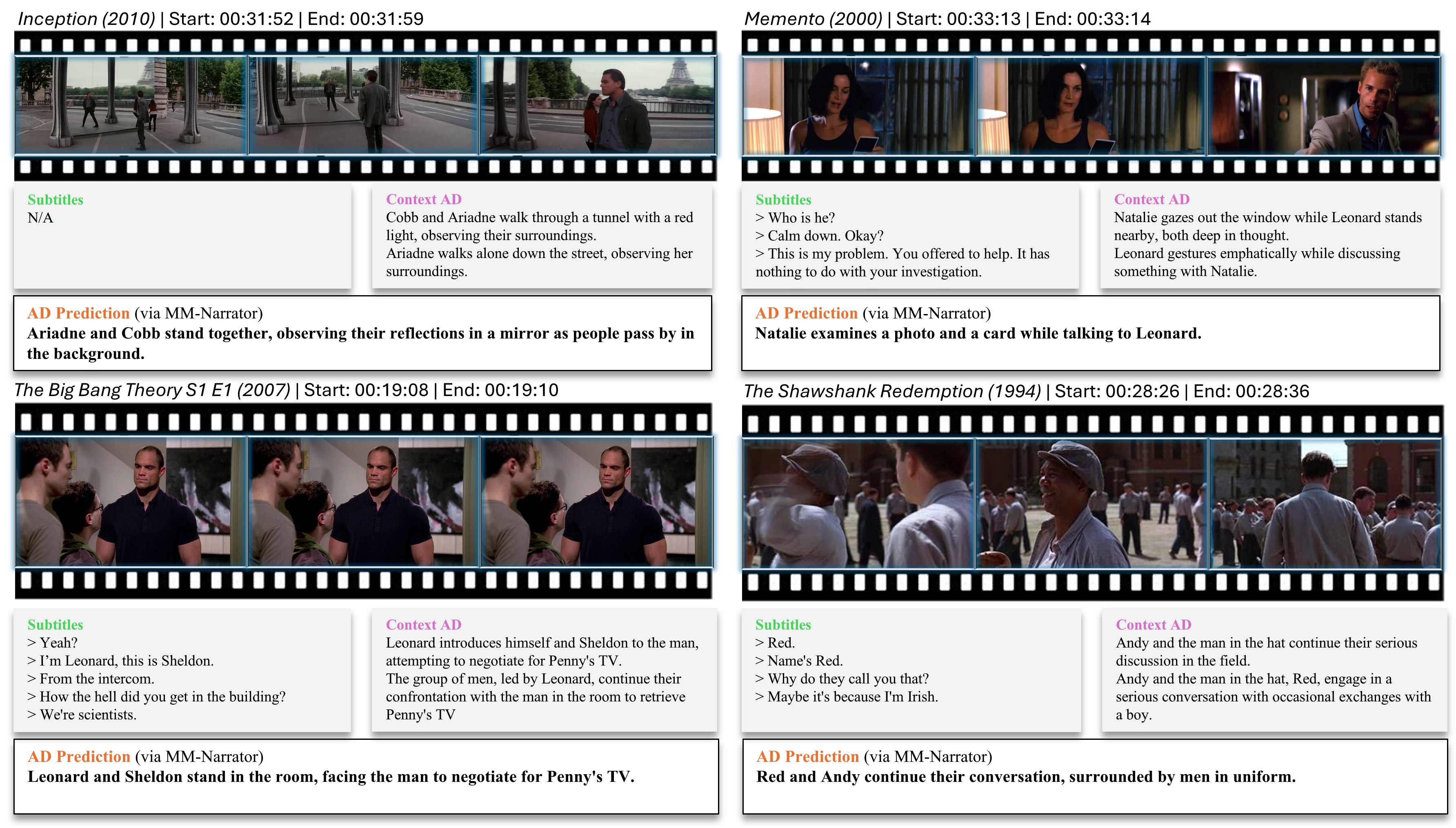}
    \caption{More qualitative demonstrations of \OURS{} on other long-form videos. For example, in \textit{Inception (2010)}, our method generates AD by conditioning on current video clip and contextual information from timestamp 00:00:00 to 00:31:52. Zoom in for details.}
    \label{fig:more_qual_extra}
\end{figure*}

\Paragraph{More qualitative results.} 
Apart from Figure ~\ref{fig:teaser} and ~\ref{fig:qual_analysis} in main paper, we show additional qualitative demonstrations of \OURS{} on both \MADEVAL{} benchmark and other long-form videos (external to the MAD-eval dataset) as Figure~\ref{fig:more_qual_dataset} and Figure~\ref{fig:more_qual_extra}, respectively, in this appendix.

\section{Discussions of Complexity-based MM-ICL}
\label{supp_sec:mm_icl}

Combining CoT with complexity-based ranking, our proposed complexity-based MM-ICL performs more favorably than classic ICL solutions. We reveal their details as follows.

\Paragraph{Reasoning with CoT.} 
We first employ GPT-4 to articulate the chain-of-thoughts (CoTs) as reasoning steps, denoted as $\mathcal{R}$, that assist in deriving the answer $\mathcal{A}$ from the question $\mathcal{Q}$.
Practically, we found a CoT-specific constraint\footnotemark{} helpful to derive reliable CoTs, ensuring a closed-loop reasoning to be inferred. 
Without this constraint, LLM might unexpectedly generate $\mathcal{R}$ followed by its own AD prediction, which are different from the human annotated $\mathcal{A}$.

\footnotetext{CoT-specific constraint: ``lets fill-in the REASONING process which derives the ANSWER from QUESTION."}

\begin{figure*}[th] 
    \centering
    \includegraphics[width=\linewidth]{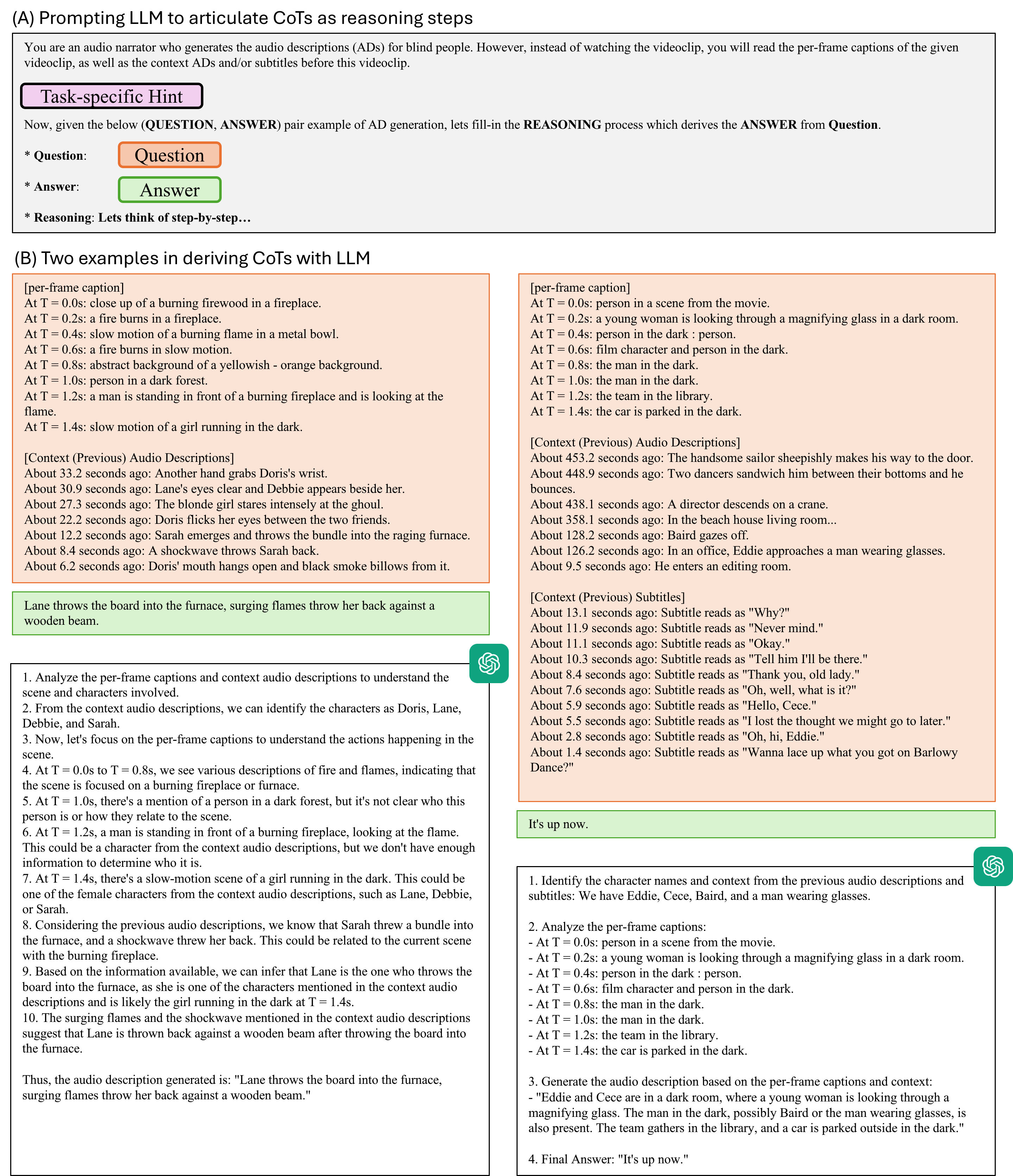}
    \caption{Reasoning with CoT. (A) An overview of prompting LLM to articulate CoTs as multimodal multi-step reasoning $\mathcal{R}$ that derives the answer $\mathcal{A}$ from question $\mathcal{Q}$, with (B) two examples shown below. Zoom in for details.}

    \label{fig:CoT}
\end{figure*}

\Paragraph{Quantifying on atomic steps.}
Practically, we observe that raw steps decided by LLM itself, might not be a considerably consistent measurement among various examples. 
Take two demonstrations shown in Figure~\ref{fig:CoT} as example: Steps 3 to 7 in \textit{left example}, conduct reasoning over per-frame captions individually, which are equivalent to step 2 in \textit{right example}, including several sub-steps in analysing the per-frame captions.
To this end, following ~\cite{fu2022complexity}, we split $\mathcal{R}$ into \textit{atomic steps} by newline char ``\texttt{\textbackslash n}", and propose using the number of atomic steps $N_\mathtt{atomic}$ as our measurement of reasoning complexity.

\begin{figure}[th] 
    \centering
    \includegraphics[width=\linewidth]{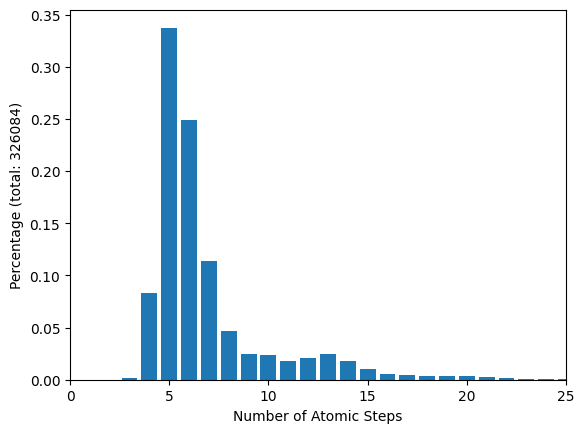}
    \caption{Distributions of multimodal \MADTRAIN{} demonstrations over reasoning complexity, quantified by $N_\mathtt{atomic}$.}
  
    \label{fig:distribution}
\end{figure}

\Paragraph{Ranking by complexity.}
We propose to select the most intuitive examples to perform few-shot MM-ICL for improving AD generation.
Here, we show the distributions of multimodal demonstrations over the complexity in Figure~\ref{fig:distribution}.
Specifically, the 10\% shortest examples lead to a simple demonstration pool $\mathcal{P}_\mathtt{simple}$ with its maximum $N_\mathtt{atomic}$ as 5, while the 10\% longest ones result in another pool $\mathcal{P}_\mathtt{hard}$ whose minimum $N_\mathtt{atomic}$ equals to 12.

\section{AD Evaluation with GPT-4}
\label{supp_sec:ad_eval}

\begin{figure*}[th] 
    \centering
    \includegraphics[width=0.95\linewidth]{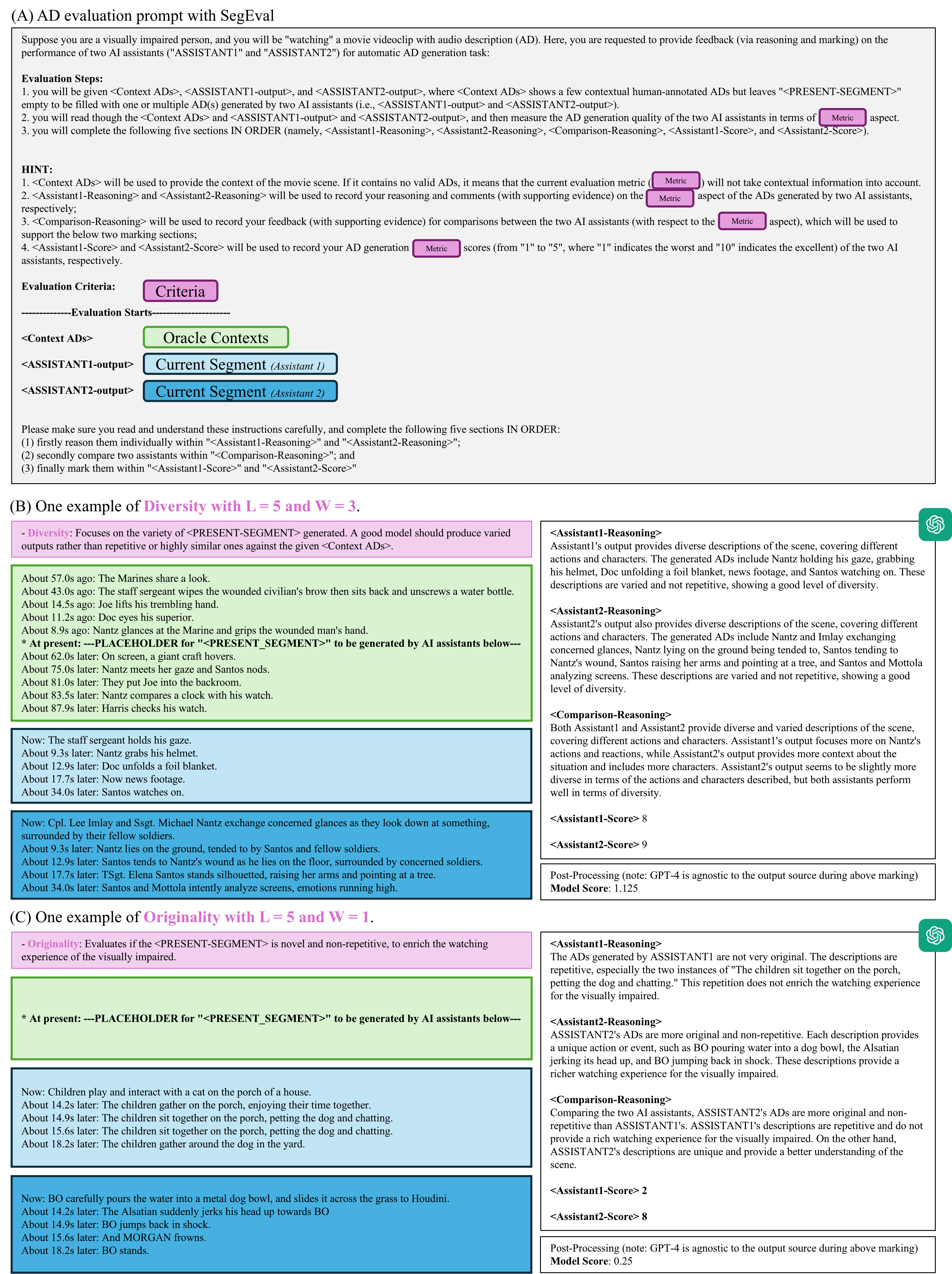}
    \caption{AD evaluation with $\mathtt{SegEval}$. (A) An overview of prompting GPT-4 to evaluate AD generation quality, with (B) one \textit{diversity} and (C) one \textit{originality} examples shown below. Zoom in for details.}

    \label{fig:eval_breakdown}
\end{figure*}

Suppose a few ADs form one segment $\mathtt{Seg}$.
For each $\mathtt{Seg}$, our proposed $\mathtt{SegEval}$ evaluator takes into consideration an oracle context window $\mathtt{Ctx}$ of length $W$, to measure its AD quality with GPT-4.
The details of $\mathtt{SegEval}$ prompt are shown as Figure~\ref{fig:eval_breakdown}.
We elaborate each individual marking criteria as follows:
\begin{itemize}
    \item \textbf{\textit{originality}}: Evaluates if the $\mathtt{Seg}$ is novel and non-repetitive, to enrich the watching experience of the visually impaired.
    \item \textbf{\textit{consistency}}: Checks if the generated $\mathtt{Seg}$ maintains a consistent tone or content throughout.
    \item \textbf{\textit{coherence}}: Determines whether $\mathtt{Seg}$ logically connects to the given $\mathtt{Ctx}$. A coherent text flows smoothly and deepen the movie understanding for the visually impaired.
    \item \textbf{\textit{diversity}}: Focuses on the variety of $\mathtt{Seg}$ generated. A good model should produce varied outputs rather than repetitive or highly similar ones against the given $\mathtt{Ctx}$.
    \item \textbf{\textit{specificity}}: Measures the level of detail in the generated $\mathtt{Seg}$, assessing if it is sufficiently detailed and/or focused for the $\mathtt{Ctx}$.
\end{itemize}
Noticeably, the first two marking aspects focus on text-level AD quality, which are context-free ($W$ = 0) evaluation metrics, while the rest three metrics measure sequence-level AD generation, taking oracle context into consideration.

\section{Implementation Details}
\label{supp_sec:impl}

\Paragraph{Multimodal Experts.} To obtain framewise image caption and people detection, we utilize vision experts publicly available via the Azure Cognitive Services APIs\footnote{\href{https://azure.microsoft.com/en-us/products/cognitive-services/vision-services}{https://azure.microsoft.com/en-us/products/cognitive-services/vision-services}}. 
For speech recognition, we choose WhisperX~\cite{bain2022whisperx} as our audio expert.
To register and recall long-term visual memory for character re-identification purpose, we adopt CLIP-ViT-L14~\cite{radford2021learning} as our visual feature extractor, and query GPT-4 as our Person-NER tool with the following prefix: \textit{``Extract the people names in the following text as a string splitted by `$|$' (return `none' if none of names are recognized): "}.

\Paragraph{Building MM-ICL Pool.}
We build the MM-ICL demonstrations for each sample in \MADTRAIN{} split~\cite{han2023autoad1}.
As the raw frames are not publicly available, we derive per-frame captions by inferring ClipCap~\cite{mokady2021clipcap} on the released CLIP-ViT features.
Differing from the main query $q_\mathtt{main}$, whose recent context ADs in $\mathcal{M}_\mathtt{short}$ are recurrently generated by \OURS{}, the queries in MM-ICL demonstrations $\mathcal{D}_{\mathtt{ICL}}$ are instead built with human annotations as their recent context ADs.
Additionally, we omit long-term visual memory retrievals when constructing MM-ICL demonstrations.

\Paragraph{GPT-4 Error Handler.}
GPT-4 might inevitably return errors when the content filtering policies\footnote{\href{https://learn.microsoft.com/en-us/azure/ai-services/openai/concepts/content-filter}{https://learn.microsoft.com/en-us/azure/ai-services/openai/concepts/content-filter}} are occasionally triggered in Azure OpenAI Service.
Such cases account for a very small proportion (less than 0.1\%), thus they would not largely affect the overall performance.
To address them, we utilize ClipCap~\cite{mokady2021clipcap} as the error handler to output video caption as AD.
Specifically, we inference ClipCap on the mean pooled feature among frames in each video clip.

\Paragraph{Deployment on Long-form Videos.}
We utilize PySceneDetect~\cite{pyscenedetect} for scene detection, and based on that, we cut long-form videos into video clips for recurrent AD generation with \OURS{}.
We utilize Google Text-to-Speech (gTTS)~\cite{gtts} for voice-over audio creation, which narrates AD for each video clip.

\Paragraph{Hyper-parameter Settings.}
Following ~\cite{han2023autoad1}, the number of frames $N$ to be sampled per video clip is set to 8, while we utilize subtitles within a time window $T_{sub}$ set to 0.25 minutes.
Our short-term memory queue is maintained to contain $K$ most recently predicted ADs with timestamps, where $K$ = 7.
The number of demonstrations $C$ equals to 5, which are sampled for conducting MM-ICL.
The API versions of GPT-4 and GPT-4V used in our experiments are `gpt4-2023-03-15' and `gpt4v-2023-08-01', respectively. 

\section{Future works}
\label{supp_sec:future}

\Paragraph{AD Generation.}
In future developments in Audio Description (AD), a critical enhancement will be the integration of advanced audio-visual speaker and character identification, coupled with strategic timing for AD delivery. 
This direction involves not only recognizing who is speaking or present in a scene but also determining the most opportune moments to provide descriptions without interrupting critical dialogue or action. 
Additionally, the establishment of a much more comprehensive and reliable external character bank, facilitating retrieval-augmented generation, will further refine AD content, ensuring it is both contextually relevant and timely. 
These advances are poised to transform AD into a more coherent, immersive experience, significantly improving accessibility for visually impaired audiences.

\Paragraph{AD Evaluation.}
In future work for AD evaluation, a crucial focus should be on enhancing the measurement of factuality, an aspect not adequately addressed by current evaluation criteria like $\mathtt{SegEval}$. 
Given the limitation of traditional reference-based scores in precisely assessing the factual accuracy of AD content, employing AI models such as GPT-4V emerges as a promising solution.
GPT-4V's advanced capabilities in understanding and contextualizing multimedia content could offer a more nuanced and accurate evaluation of AD factuality. 
This shift towards AI-driven, factuality-focused evaluation methods would not only provide a more comprehensive assessment of AD quality but also ensure that the generated descriptions are reliably accurate, ultimately benefiting visually impaired individuals with a more authentic storytelling experience.


\end{document}